# Philosophy-informed Machine Learning

M.Z. Naser, PhD, PE

Artificial Intelligence Research Institute for Science and Engineering (AIRISE), Clemson University, Clemson, SC, USA

E-mail: mznaser@clemson.edu, Website: www.mznaser.com

## Abstract

Philosophy-informed machine learning (PhIML) directly infuses core ideas from analytic philosophy into ML model architectures, objectives, and evaluation protocols. Therefore, PhIML promises new capabilities through models that respect philosophical concepts and values by design. From this lens, this paper reviews conceptual foundations to demonstrate philosophical gains and alignment. In addition, we present case studies on how ML users/designers can adopt PhIML as an agnostic post-hoc tool or intrinsically build it into ML model architectures. Finally, this paper sheds light on open technical barriers alongside philosophical, practical, and governance challenges and outlines a research roadmap toward safe, philosophy-aware, and ethically responsible PhIML.

*Keywords:* Engineering, Artificial intelligence, Philosophy.

## 1.0 Introduction

Contemporary machine learning (ML) systems have been noted to achieve notable performance on defined tasks yet fail catastrophically when deployed in contexts that demand genuine understanding [1]. A deep dive into the open literature shows that there are three fundamental limitations to current ML approaches, namely *blackbox brittleness* (which renders models uninterpretable and unreliable under distribution shift [2]), *causal blindness* (which conflates correlation with causation [3]), and *alignment failures* (which produce systems optimizing objectives misaligned with human values [4]). These deficiencies stem from a profound philosophical poverty in how ML conceptualizes knowledge, reasoning, and values.

The first fundamental limitation, blackbox brittleness, manifests when trained models fail on seemingly trivial variations of their training distribution. For example, a vision model that accurately identifies stop signs under normal conditions might misclassify them entirely when small adversarial perturbations are applied [5]. Not surprisingly, the same brittleness extends beyond adversarial examples to everyday distribution shifts (e.g., natural language processing models exhibit performance degradation when processing text from different cultural contexts, etc.) [6]. This behavior reveals that the model has learned superficial statistical regularities rather than robust conceptual understanding [7]. The opacity of these failures and our inability to understand why a model made specific errors compound the problem of predicting/preventing future failures through traditional approaches [8].

Causal blindness represents an even more fundamental limitation, as ML models typically learn associations without understanding the causal mechanisms that generate observed data [9]. For example, a model might learn that specific genetic markers correlate with disease outcomes without understanding whether these markers cause the disease, are caused by it, or merely share common causes [10]. This limitation becomes critical when such models are used to inform interventions (e.g., a healthcare system that recommends treatments based on correlational patterns might suggest ineffective/harmful interventions because it cannot distinguish causal relationships from spurious associations) [11]. Such an inability to reason severely limits the applicability of ML in domains where understanding causal mechanisms is essential for effective action [12].

The third fundamental limitation pertains to alignment failures. Here, the specification of reward/loss functions that truly capture human preferences proves difficult and often leads to systems that exploit loopholes or optimize for easily measurable proxies (rather than intended outcomes) [13]. This can be seen in recommender systems that optimize for engagement metrics that promote addictive usage patterns, etc. [14]. Such alignment failures reflect technical limitations and philosophical challenges in formalizing human values and ensuring that ML systems respect them [15].

Philosophy-informed ML (PhIML) addresses these foundational challenges by directly integrating analytic philosophy into the design of ML systems. Simply, rather than treating philosophy as an abstract complement to technical work, PhIML embeds philosophical principles into model architectures and training/evaluation procedures. The result is a new class of ML systems that exhibit qualitatively different capabilities. PhIML has broader roots that start with early logicist ML, exemplified by theorem provers, which attempted to encode human knowledge in formal logical systems but struggled with the difficulty of hand-crafted rules and uncertainty [16]. Then, the connectionist revolution succeeded in pattern recognition at the cost of sacrificing interpretability and reasoning capabilities. The emergence of attempts to combine neural learning with symbolic reasoning is promising, but often lacks principled foundations for this integration [17].

More recently, the rise of -*informed* research across ML reflects growing recognition that purely data-driven approaches have central limitations. For example, physics-informed neural networks (PINNs) incorporate physical laws as inductive biases to improve scientific applications' generalization and interpretability [18,19]. Similarly, biology-informed models integrate evolutionary principles and biological constraints to improve performance on tasks such as gene research [20]. Further, cognitively-informed approaches draw on human cognitive architecture to design more sample-efficient and robust learning algorithms [21]. As one can see, PhIML represents a natural extension of this trend by recognizing that philosophy (i.e., a discipline that has systematically studied knowledge, reasoning, causation, and values) offers essential insights for addressing the foundational challenges facing ML [22].

The scope of this work encompasses four major areas where philosophy and ML productively intersect. The first three sections outline a conceptual look at PhIML, wherein Section 2.0 examines how epistemological theories and logic enhance uncertainty quantification and knowledge representation in learning systems, Section 3.0 explores how philosophical theories of causation enable ML models to reason about interventions and counterfactuals, and Section 4.0 investigates how moral philosophy informs the design of value-aligned systems that respect human values and ethical constraints. Each of these sections showcases the application of PhIML via conceptual case studies. Then, we move to Section 5.0, where we conclude with a discussion of key challenges and future outlooks.

## 2.0 Philosophical foundations of epistemology & logic

The philosophical foundations of PhIML rest on four pillars drawn from analytic philosophy: i) epistemology's treatment of knowledge and uncertainty, ii) formal logic's for reasoning, iii) theories of causation, and iv) ethics' approaches to value and right action. This section examines the first two pillars to demonstrate how integrating philosophical insights can improve model architectures and training procedures at a theoretical and philosophical level[1].

### *2.1 Epistemology & uncertainty*

Traditional ML approaches rely almost exclusively on probability theory to represent uncertainty through probability distributions over outcomes [23]. While computationally tractable, this probabilistic framework collapses important epistemic distinctions that philosophers have long recognized. When a ML model assigns high probability to a prediction, it provides no mechanism for distinguishing whether this confidence stems from robust evidence, spurious training correlations, or systematic biases in data collection [3]. This issue could be addressed by adopting philosophy-informed approaches to incorporate richer representations of epistemic states that preserve the aforementioned distinctions. The discussion that follows builds from this critique toward progressively richer representational remedies.

The tension between justified true belief and Bayesian belief reflects deeper philosophical debates about the nature of rational belief [24]. On the one hand, justified true belief requires binary judgments—either one knows something, or one does not. On the other hand, Bayesian approaches assign graded confidence to propositions [25]. Traditional epistemology's emphasis on justification highlights what probabilistic approaches often mask: the importance of understanding why we believe what we believe. Take the following example: a ML system that predicts disease with 90% confidence based on spurious correlations provides the same output as one that reaches an identical confidence through causally grounded reasoning, yet these two systems differ fundamentally in their reliability and interpretability [26].

Rather than representing epistemic states through single probability distributions, credal sets can employ convex sets of probability distributions to capture forms of uncertainty that single distributions cannot express [27]. This approach aligns with the philosophical recognition that rational agents need not always possess precise probabilistic beliefs, as sometimes a given evidence supports a range of probability assignments (rather than a unique distribution) [28]. Thus, in ML applications, credal sets enable models to represent the difference between risk (where probabilities are well-defined) and ambiguity (where the appropriate probability model itself remains uncertain). It should be noted that the implementation of credal sets in neural architectures requires some reimagining of standard training procedures to minimize the volume of credal sets while ensuring they contain the true distribution to balance precision with robustness to model misspecification [29]. As one can see, credal sets address only quantitative facets of uncertainty, and in order to capture the structural provenance of beliefs, one must adopt logical formalisms.

---

[1] It should be noted that the selected ML models were chosen for their simplicity and wide use. These were also used in their default settings to enable direct replication. Further, the first two case studies present a collection of philosophical concepts in terms of PhIML when applied as post-hoc or intrinsic methods. Then, in case study no. 3, only one concepts is applied at both the post-hoc stage, as well as an intrinsic method. Two of the three case studies were on classification problems, and one case study was on regression. This rationale was adopted here for brevity and to showcase the use of various possible uses of PhIML. Therefore, thorough treatment for each philosophical concept or equivalent computational treatment was avoided here to avoid pushing this paper away from its theme. Interested readers are invited to adopt the above rationale or develop new/elegant approaches. The author hopes that his future work will be focused on specific treatments for particular concepts/ML models.

Epistemic logic extends uncertainty representation beyond probabilistic frameworks by providing formal languages for reasoning about knowledge and belief [30]. The modal operators of epistemic logic—knowing $p$, believing $p$, and their iterations—capture patterns of reasoning essential for multi-agent learning systems [31]. When multiple models coordinate predictions or share learned representations, epistemic logic offers a principled approach for aggregating partial knowledge while tracking the provenance and reliability of different information sources [32]. The axioms of epistemic logic, such as the *knowledge axiom* (what is known is true) and *positive introspection* (if one knows $p$, one knows that one knows $p$), translate into architectural constraints that ensure coherent reasoning about uncertainty. This logical layer can also benefit from knowledge graphs with epistemic operators to demonstrate the practical value of these philosophical frameworks.

Traditional knowledge graphs represent facts as edges between entities, but epistemic knowledge graphs augment these representations with modal qualifiers tracking the epistemic status of each relation [33]. For example, a financial knowledge graph might encode not just that two companies have a supplier relationship, but which agents know this relationship, with what confidence, based on what evidence, and under what conditions this knowledge remains valid. This richer representation enables sophisticated reasoning about incomplete and potentially conflicting information. Figure 1 shows a simple example of how traditional knowledge graphs encode relationships between hypothetical entities as simple edges (e.g., *ChipMaker Inc supplies TechCorp.*). However, the companion epistemic knowledge graphs augment these base relations with modal qualifiers that capture the knowledge state of different agents. In the epistemic representation, the same supplier relationship is enriched with metadata indicating that a *Supply Analyst* knows this fact with 85% confidence based on shipping data, while an *AI Monitor* inferred it with only 60% confidence from SEC filings. This additional epistemic layer enables systems to reason about information uncertainty, source reliability, and knowledge provenance.

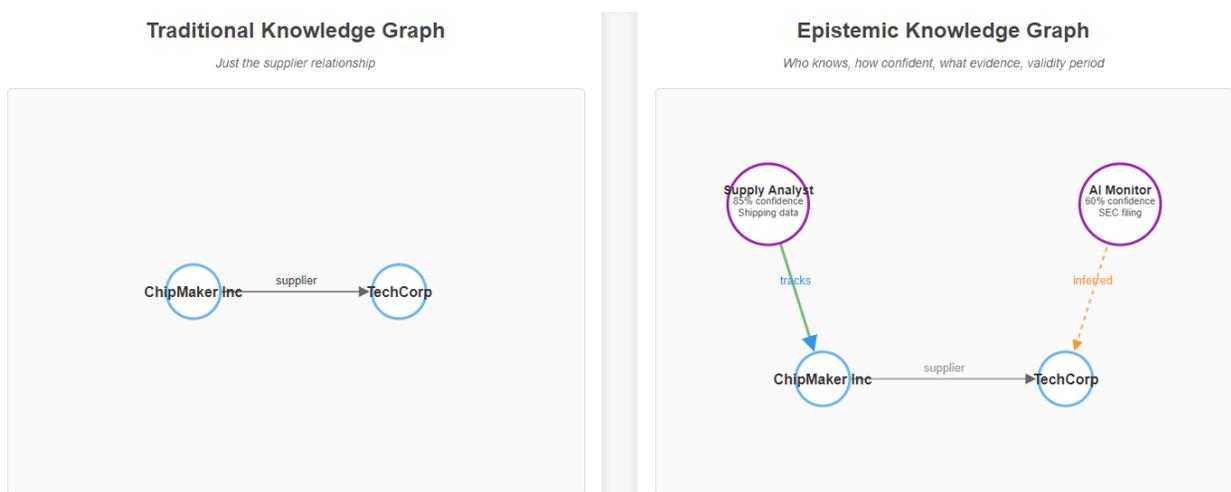

Fig. 1 Traditional vs. epistemic graphs

## 2.2 Formal logic for ML

Formal logic provides the second philosophical pillar for PhIML and offers precise languages for expressing constraints, reasoning patterns, and safety requirements that pure ML approaches struggle to capture [34]. While early ML extensively employed logic, the integration of logical frameworks with modern algorithms requires new approaches that preserve the flexibility of traditional algorithms while gaining the guarantees that formal systems provide (e.g., first-order logic constraints enable the specification of safety requirements that must hold universally, modal/temporal logics capture sequential/possibility-based reasoning needed in planning and verification, and paraconsistent logics provide frameworks for learning from inconsistent or noisy data without logical explosion) [35]. From this lens, the discussion first turns to first-order logic (FOL), whose universal and existential quantifiers make it the most direct bridge between classical symbolic guarantees and ML [36].

More specifically, FOL constraints can be seen in critical systems, which require guarantees that certain conditions never occur (e.g., a financial system must never violate regulatory requirements, etc.). Expressing these requirements in FOL provides precision that natural language specifications lack [37]. For example, logic-regularized networks can add penalty terms to the loss function that measure violation of logical constraints (with careful scheduling that gradually increases the penalty weight to ensure eventual satisfaction while maintaining training stability) [38]. Similarly, semantic loss functions can directly encode logical formulas into differentiable objectives using fuzzy logic semantics (to enable end-to-end training of networks that satisfy complex specifications). Another example can be seen in projection-based methods, which can alternate between unconstrained gradient steps and projection onto the feasible region defined by logical constraints to guarantee the satisfaction throughout training while exploring the space of compliant models [39].

With the above static constraints operationalized, the focus naturally shifts to time-dependent possibilities. Accordingly, modal and temporal logics address the critical need for reasoning about possibilities, necessities, and temporal evolution in sequential decision-making tasks [40]. Here, standard neural networks may struggle with naturally expressed tasks in modal logic, such as tasks requiring reasoning about what might happen or what must eventually occur [41]. Thus, temporal logic provides operators for expressing properties like *eventually reach the goal* or *always avoid obstacles* that can be essential for specifying objectives in planning/control applications [42]. To turn these abstract advantages into models, researchers can design neural architectures that explicitly mirror the semantics of modal logic. For instance, implementations of modal logic could employ architectures that mirror the possible-worlds semantics underlying modal reasoning, where each possible world corresponds to a latent representation. Modal operators like *necessity* and *possibility* become operations on these structured representations, with backpropagation flowing through the logical structure [43]. This approach enables models to learn which possibilities are relevant for particular tasks while respecting the logical constraints that modal operators impose. Yet, even architectures equipped with temporal reasoning must confront contradictory and noisy data—an issue best met by paraconsistent logic [44].

Here, paraconsistent logics can help modify classical inference rules to localize contradictions and prevent them from contaminating all reasoning [45]. In ML contexts, paraconsistent shields filter training data to identify and quarantine contradictory examples while preserving consistent information [36]. These shields can simply employ relevance logic, which tracks the actual use of premises in derivations to ensure that contradictions only affect conclusions that depend on inconsistent data. For example, training on noisy labels can benefit from paraconsistent preprocessing that identifies label contradictions without discarding entire training examples [46]. Another example that naturally fits herein is reasoning with disagreement, which can be seen in multi-annotator scenarios, where different experts provide conflicting labels. Thus, one can develop neural architectures that internally implement paraconsistent reasoning that utilize specialized activation functions that behave like paraconsistent logical operators and do not undermine other classes of logical reasoning [47].

### *2.3 Case study no. 1*

This experiment evaluates how integrating philosophical principles from epistemology and formal logic improves model behavior when standard classifiers (random forests (RF) and support vector machines (SVMs) when used in their default settings from the Scikit-learn package) encounter scenarios requiring adherence to domain ontologies. We present the merit of PhIML as a post-hoc tool, as well as an intrinsic process via four controlled scenarios: 1) legal document classification with mutual exclusivity between contracts and patents, 2) medical diagnosis respecting severity hierarchies (*Pneumonia→Flu→Mild Symptoms*), 3) constraint-aware training that penalizes logical violations during learning, and 4) a logic-guided architecture combining multiple constraint types. Here we discuss each scenario (see data distributions in Fig. 2):

- **Scenario 1 (post-hoc): Document classification (Contract ⊥ Patent)** directly implements the first-order logic constraints discussed in Sec. 2.2, where ML systems require guarantees that certain conditions never occur. The mutual exclusion constraint is formally expressed as:

$$\forall x: \neg(Contract(x) \land Patent(x)) \tag{1}$$

  to prevent the logical contradiction of a document being simultaneously classified as both contract and patent. This post-hoc enforcement mechanism operationalizes this through probability adjustment:

$$\text{when } P(Contract|x) > \tau \land P(Patent|x) > \tau, \tag{2}$$

  the system reduces the lower probability by an arbitrarily selected factor of 0.3 to project the prediction onto the feasible region defined by the logical constraint. This mirrors the *projection-based methods* that alternate between unconstrained gradient steps and projection onto the feasible region defined by logical constraints to ensure that licensing agreements with mixed terminology do not violate domain ontology despite statistical similarities to both categories.

- **Scenario 2 (post-hoc): Severity hierarchy (pneumonia → Flu → Mild)** exemplifies the modal logic principles also presented in Sec. 2.2, particularly with regard to reasoning about what might happen or what must eventually occur. The hierarchical constraints are expressed using the necessity operator:

$$\Box(Pneumonia(x) \rightarrow Flu(x)) \land \Box(Flu(x) \rightarrow Mild(x)), \tag{4}$$

which encodes medical knowledge that severe conditions necessarily imply milder symptoms. This implementation uses probability transfer mechanics where

$$\text{if } P(Pneumonia|x) > \tau \wedge P(Flu|x) < \tau, \tag{5}$$

the system calculates:

$$\delta = min(P(Pneumonia|x) \times 0.3, \tau - P(Flu|x)) \tag{6}$$

and updates:

$$P'(Flu|x) = P(Flu|x) + \delta. \tag{7}$$

This ensures epistemic coherence by preventing the model from asserting high confidence in pneumonia while denying flu symptoms. The same also addresses the epistemological concern that purely probabilistic approaches cannot adequately capture the distinction between correlation and genuine medical causation.

- **Scenario 3 (intrinsic/during training): Constraint-aware loss function** demonstrates the logic-regularized networks approach discussed in Sec. 2.2 that adds penalty terms to the loss function that measure violation of logical constraints. Here, the augmented loss function:

$$L_{total} = L_{base} + \lambda \sum_i V(x_i) \tag{8}$$

incorporates violation scores $V(x_i) = min(P(Flu|x_i), P(Pneumonia|x_i))$ when both probabilities exceed threshold $\tau$. This directly encodes the mutual exclusion constraint into the training objective. In addition, a progressive weight adjustment $w_i = exp(-\alpha \times V(x_i))$ implements careful scheduling that gradually increases the penalty weight to ensure eventual satisfaction while maintaining training stability. This approach enables the model to learn representations that inherently respect logical constraints rather than requiring post-hoc correction, thereby addressing the fundamental epistemological issue that traditional ML provides no mechanism for distinguishing whether confidence stems from robust evidence, spurious training correlations, or systematic biases.

- **Scenario 4 (intrinsic/during training): Logic-guided architecture** represents a sophisticated integration of philosophical principles by implementing an explicit logical reasoning layer that mirrors the possible-worlds semantics underlying modal reasoning discussed in Sec. 2.1. In this scenario, the architecture:

$$h = f_{base}(x) \rightarrow P_{base} \rightarrow LogicLayer(P_{base, \ constraints}) \rightarrow P_{final} \tag{9}$$

creates a neural structure where modal operators like necessity and possibility become operations on these structured representations. Such a logic layer simultaneously handles mutual exclusion $(P(A) > \tau \wedge P(B) > \tau \rightarrow$ reduce both) and implication enforcement $(P(A) > \tau \wedge P(B) < \tau \rightarrow transfer \ A \rightarrow B)$. This demonstrates how epistemic logic offers a principled approach for aggregating partial knowledge while tracking the provenance and reliability of different information sources. This architectural approach ensures that backpropagation flows through the logical structure and enables the model to learn which possibilities are relevant while respecting the logical constraints that modal operators impose, effectively bridging the gap between statistical learning and formal reasoning.

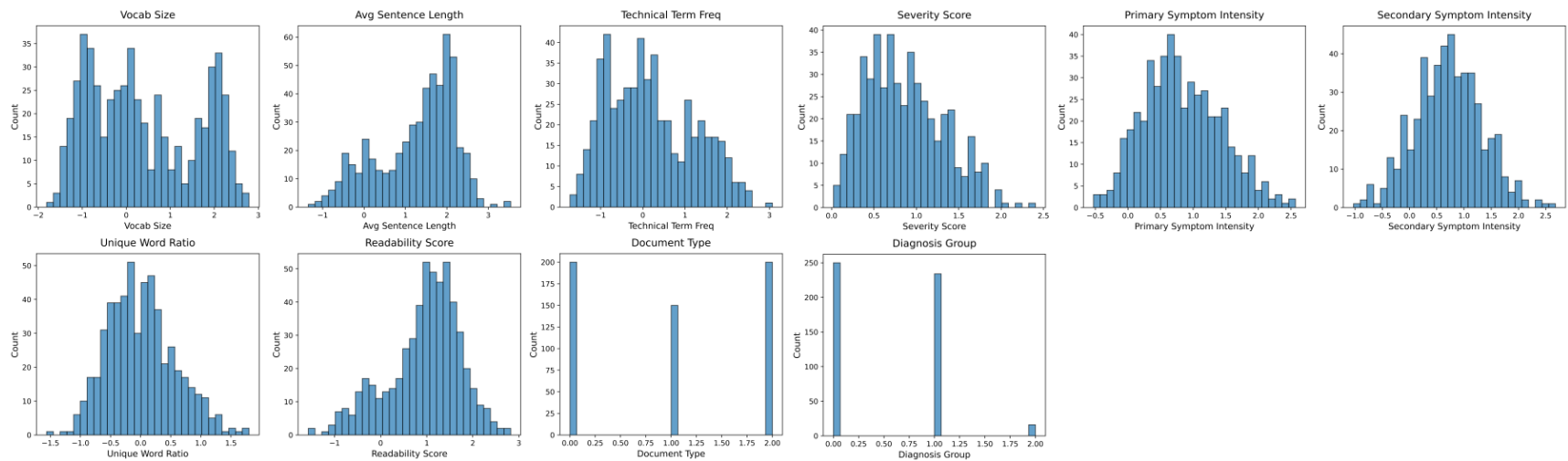

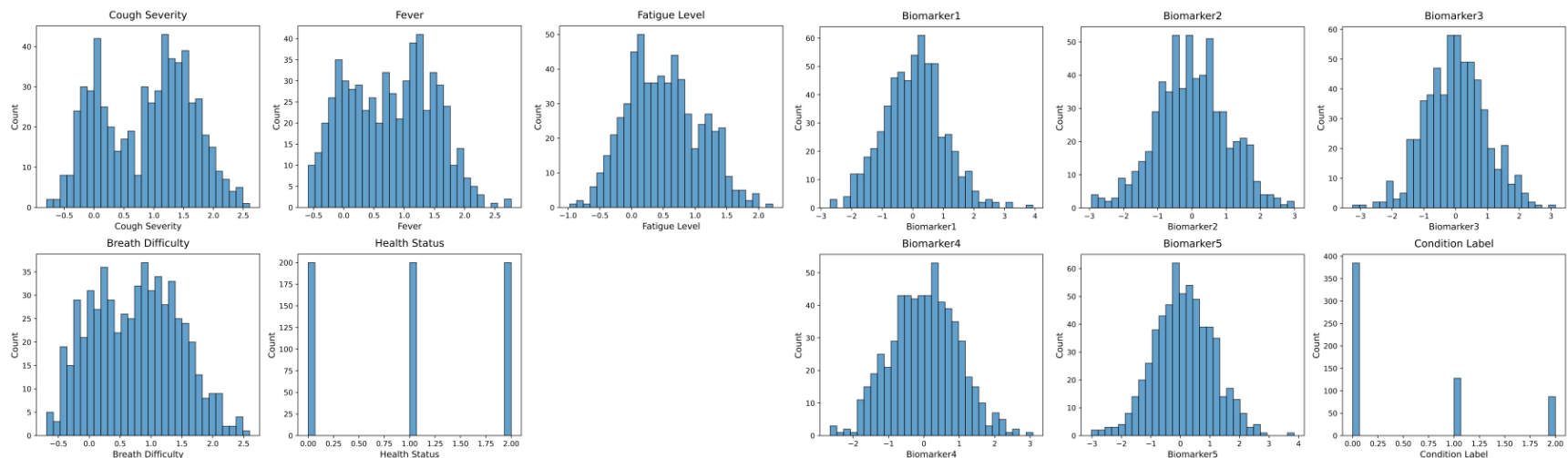

Fig. 2 Data and features used in case study no. 1

Figure 3 presents the results from the above-described experiments and demonstrates PhIML's effectiveness in reducing logical constraint violations while preserving predictive accuracy across different algorithms and scenarios. More specifically, post-hoc enforcement achieves the most dramatic violation reductions, with *Document Classification* eliminating violations entirely (13.3%→0% for RF, 8.5%→0% for SVM) and *Severity Hierarchy* reducing violations by 76% for RF (36%→8.7%) and 68% for SVM (42%→13.3%), all while maintaining identical accuracy levels. This is expected: the post-hoc repair is meant to improve logical coherence, not factual accuracy. In contrast, the intrinsic approaches show more nuanced trade-offs with *Constraint-Aware Loss* achieves modest violation reductions (approximately 2-6 percentage points) with minimal accuracy impact, while *Logic-Guided Architecture* delivers strong violation reduction (73% for RF, 43% for SVM) and even improves SVM accuracy by 1.7 percentage points. Overall, and across all eight experiments, accuracy changes range from -0.6% to +1.7%, with most scenarios showing zero or positive impact.

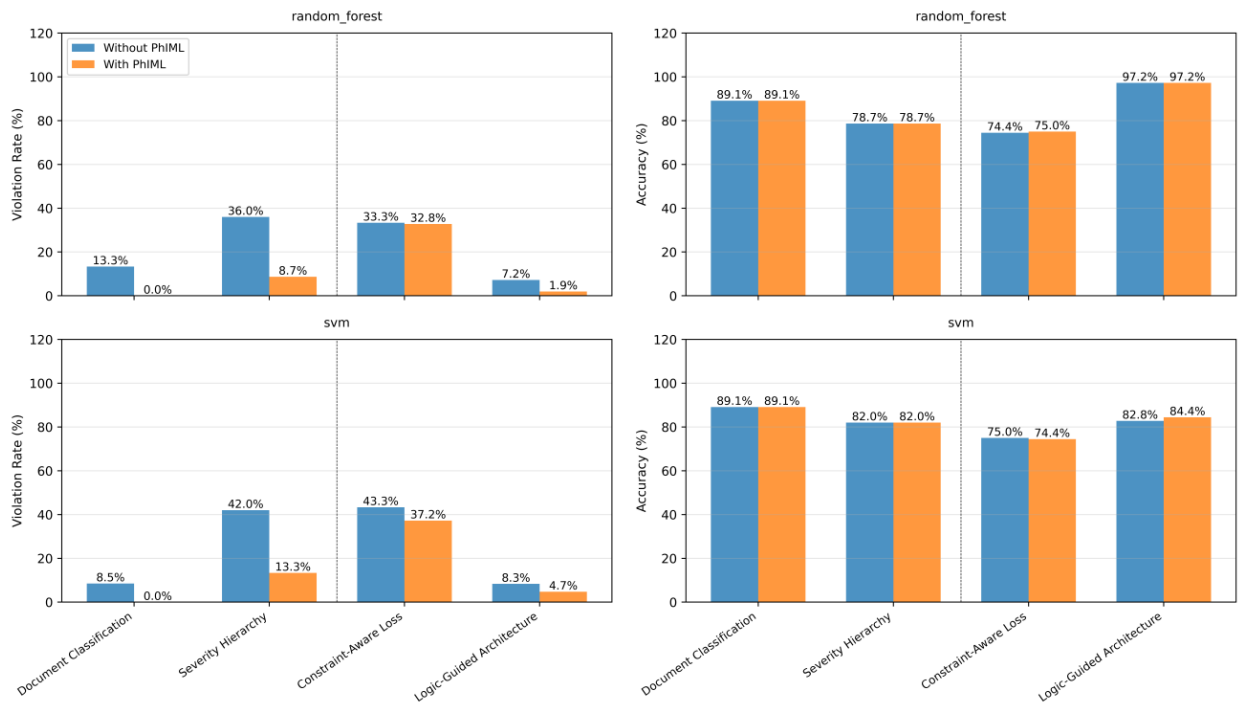

Fig. 3 Results on case study no. 1

## 3.0 Causation theories
The philosophy of science continues to face the ongoing challenge of understanding causal relationships. This has led to the development of frameworks for causal reasoning that PhIML can translate into computational methods, as seen below.

### *3.1 Counterfactual theories & causal frameworks*
Counterfactual theories revolve around the idea that causes are factors whose modification would change outcomes. But, the integration of causal inference requires careful attention to philosophical subtleties because not all counterfactuals are causal, and the relationship between counterfactual dependence and causation requires nuanced treatment. Such a treatment can be directly translated into computational frameworks through potential outcome models and structural causal graphs. In the potential outcomes, each unit is associated with multiple potential responses under different treatments, with causal effects defined as differences between these potential outcomes. This directly implements the philosophical idea that causation involves comparison across possible worlds (i.e., what would have happened under different circumstances?). Yet, we can observe only one potential outcome for each unit, which constitutes a key philosophical challenge to empirically access counterfactual worlds. This also motivates statistical techniques such as propensity-score matching and inverse-probability weighting.

Building on the above, deep causal graphs generalize traditional directed acyclic graphs (DAGs) by incorporating latent variables, nonlinear relationships, and uncertainty about graph structure. While standard causal graphs assume fixed, known structures, deep causal graphs learn hierarchical representations where structure and parameters undergo joint optimization. This approach aligns with philosophical views of causation as multi-level, wherein micro-level mechanisms give rise to macro-level causal relationships. Consequently, ML models can learn representations at multiple scales, and when supplemented with attention mechanisms, can also discover causal connections. The architectural consequences of these representations become most evident when the principle of minimal change is made explicit.

The philosophical principle of *minimal change* (i.e., counterfactuals should differ minimally from actuality except as required by the antecedent) can be operationalized via regularization terms that encourage sparse modifications. Here, model architecture can incorporate separate pathways for factual and counterfactual processing, with shared representations ensuring consistency while specialized components handle world-specific reasoning. This layered design completes the progression from philosophical counterfactual semantics to trainable models that yield systems that respect causal logic while retaining the adaptability of deep learning.

### 3.2 Interventionism & causal invariance

Woodward's interventionist account of causation provides an alternative philosophical foundation [48]. Unlike counterfactual theories, interventionism defines causation through hypothetical interventions: $X$ causes $Y$ if intervening on $X$ would change $Y$ [49]. Simply, adopting the principle that causal relationships remain invariant under intervention, while mere correlations break down, provides both a definitional criterion and a discovery method for causal structure. This computationally translates into Pearl's *do*-calculus (where the *do*-operator formally represents interventions that break natural causal relationships) to deliver algorithms for computing intervention effects from observational data [50]. For example, computing $do(X=x)$ requires techniques different from conditioning on $X=x$, as interventions eliminate confounding paths that conditioning preserves.

The above-mentioned philosophy-informed invariance principles can be used to guide the development of causal representation learning by augmenting standard neural networks with intervention-modeling capabilities [51]. Here, the philosophical claim that causal mechanisms operate independently can be translated into the independent causal mechanisms principle and implemented through modular neural architectures where components can be modified without affecting others. For example, special training procedures can be adopted to encourage invariance by presenting data from multiple environments, with causal features identified as those whose relationships remain stable across contexts [52]. This approach succeeds where purely statistical methods fail because spurious correlations that happen to hold across training environments can be easily distinguished from genuine causal relationships through their behavior under intervention [53].

One should still note that the practical implementation of interventionist principles confronts the challenge that real interventions are often impossible or unethical [54]. On a more positive note, philosophy-informed approaches may address this through techniques for causal discovery from observational data. For instance, natural experiments, instrumental variables, and regression discontinuity designs, which happen to be motivated by interventionist thinking, can allow causal inference without true randomization [55]. Therefore, possible neural implementations can learn to recognize and exploit these quasi-experimental variations, with architectures designed to separate variation due to intervention-like natural occurrences from confounded observational correlations.

### *3.2 Case study no. 2*

In order to demonstrate how PhIML can improve both post-hoc predictions and in-training invariance, we construct two illustrative scenarios and use two ML algorithms (RF and a simple linear model). The first (scenario no. 5) shows how a trained model's counterfactual predictions can be *too wild* and violate a minimal-change principle (as discussed in Sec. 3.1); we then apply a PhIML enforcer to repair these predictions. The second (scenario no. 6) shows an interventionist or invariance-driven design as discussed in Sec. 3.2 of a pooled model across multiple environments (domains), and how a lightweight PhIML can shrink that performance gap by explicitly modeling environment-specific experts and blending them with an environment identifier. In detail:

- **Scenario 5 (post-hoc): counterfactual consistency**. Let each example be $(x_i, t_i, y_i)$, where $x_i \in R^3$ (age, severity, comorbidity), $t_i \in \{0,1,2\}$ is the observed treatment, and $y_i \in R$ the continuous outcome – see Fig. 4. We train a regressor model $f$ on the augmented inputs *(x,t)*. For any test point $i$, define the factual prediction $\hat{y}_i^{\text{factual}} = f(x_i, t_i)$, and for each alternative treatment $t' \neq t_i$ the counterfactual prediction $\hat{y}_i^{\text{cf}} = f(x_i, t')$. Then, a minimal-change enforcer with threshold $\tau > 0$ then produces

$$\widehat{y_{i,t'}}^{\text{adj}} = \begin{cases} \widehat{y_{i,t'}}^{\text{cf}}, & \left|\widehat{y_{i,t'}}^{\text{cf}} - \hat{y}_i^{\text{factual}}\right| \leq \tau \\ \hat{y}_i^{\text{factual}} + sign\left(\widehat{y_{i,t'}}^{\text{cf}} - \hat{y}_i^{\text{factual}}\right)\tau, & otherwise \end{cases} \tag{10}$$

We then reassess all adjusted predictions under the same rule. Therefore, the violation can be calculated such that for each test example *i*, define a binary indicator:

$$v_i = 1\left(\max_{t' \neq t_i} \left|\widehat{y_{i,t'}^{\text{cf}}} - y_i^{\widehat{\text{factual}}}\right| > \tau\right) \tag{11}$$

And, the overall violation rate is:

$$\text{Violation} = \frac{1}{N}\sum_{i=1}^{N} v_i \tag{12}$$

- **Scenario 6 (intrinsic/during training): environment ensemble.** Suppose we have four environments $e = 0,1,2,3$, each with its own distribution $P_e(X,Y)$ – see Fig. 4. We hold out $E_{\text{train}} = \{0,1\}$ for training and $E_{\text{test}} = \{2,3\}$ for evaluation. For each $e \in E_{\text{train}}$, we fit a regressor $g_e(x)$. We then build a meta-model $h(g_0(x), g_1(x), e')$  where $e' \in \{0,1,2,3\}$ is a numeric environment identifier. The meta-model learns to weight expert outputs plus the ID feature. At test time in an unseen environment $e \in \{2,3\}$, we predict $\hat{y} = h(g_0(x), g_1(x), e)$. That final integer $e_{test} \in \{2,3\}$ was *never seen* in training, so the model is forced to extrapolate along, exactly matching the intended *environment ensemble* idea (this feature acts like a causal instrument). Empirically: The violation in this scenario can be calculated by computing the mean square error (MSE) in each held-out environment $e$: $\text{MSE}_e = E_{(x,y) \sim P_e}[(y - \hat{y})^2]$ and then, the violation is the variance across these MSEs: $\text{Violation} = \text{Var}(\{\text{MSE}_e\})$ and the mean squared error over all test points (or pooled test environments) $\text{MSE} = \frac{1}{N}\sum_{i=1}^{N}(y_i - \hat{y}_i)^2$.

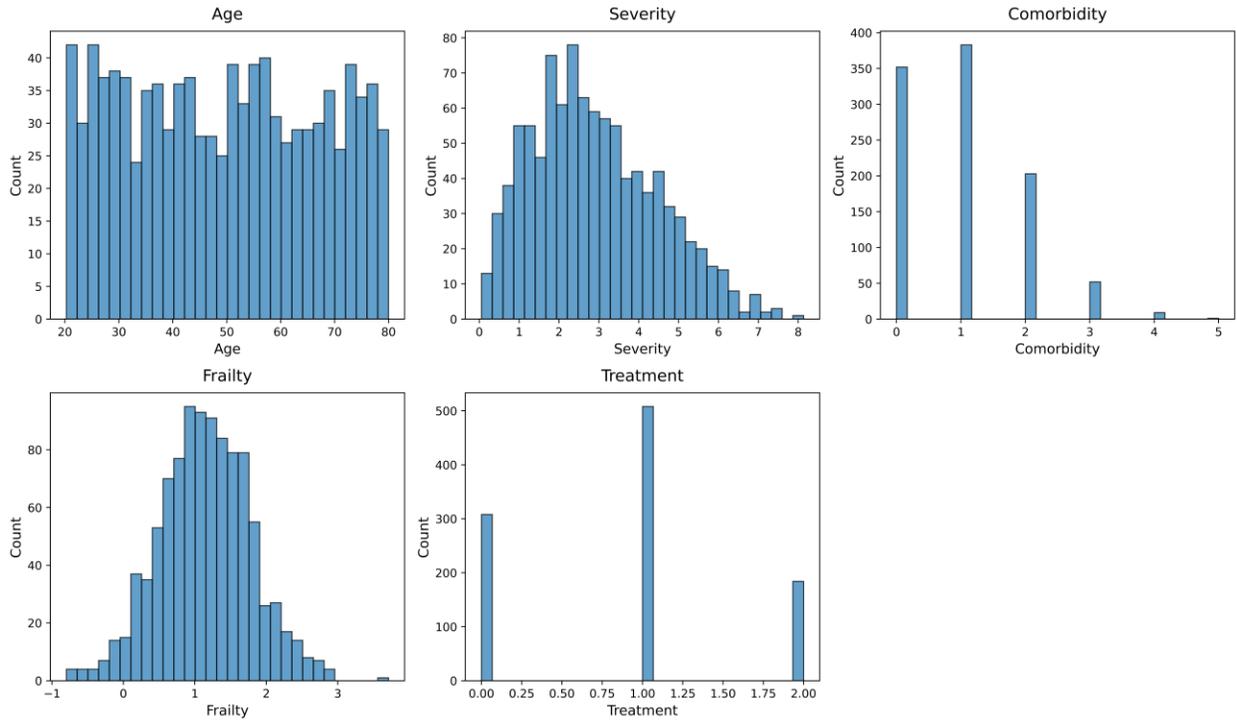

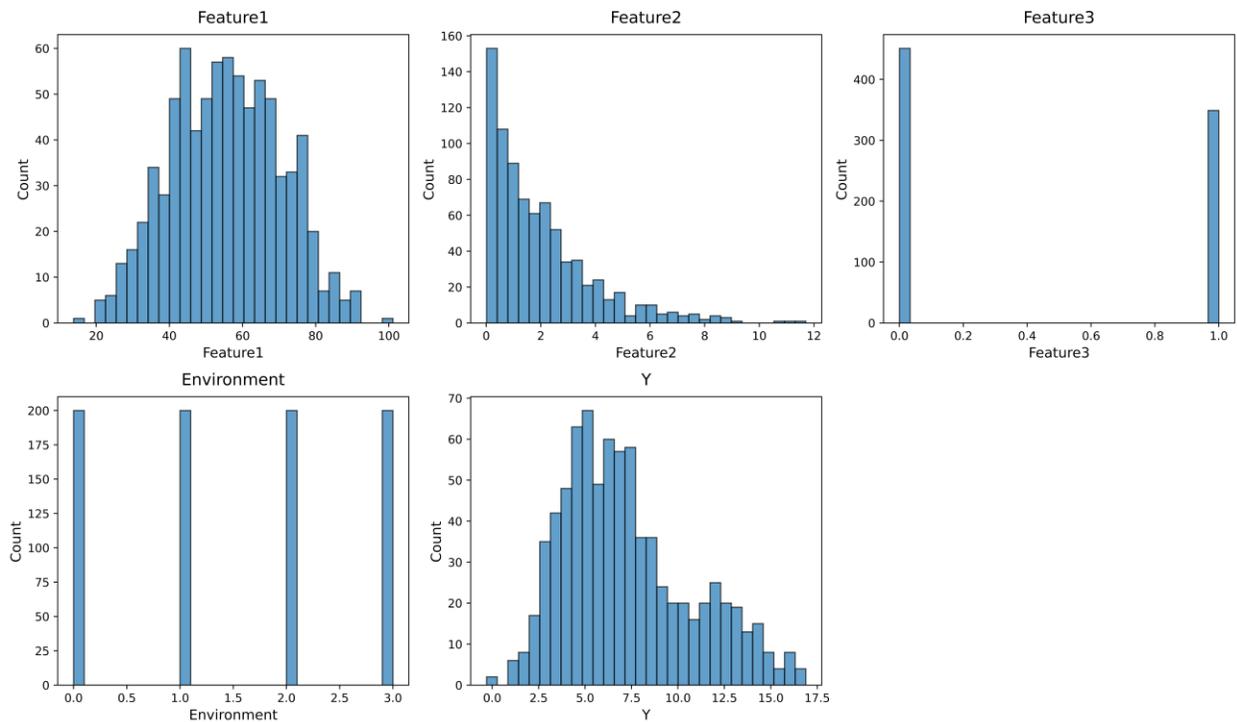

Fig. 4 Data and features used in case study no. 2

Across both model classes, enforcing PhIML constraints yields a reduction in violations with minimal/beneficial impact on overall fit (see Fig. 5). More specifically, for the RF regressor in the counterfactual consistency scenario, the violation rate falls from 38.3% to 3.0%, while the MSE remains unchanged at 0.6709. In the environment ensemble setting, variance in per-environment error drops from 19.86 to 7.51, accompanied by a reduction in mean MSE from 8.97 down to 6.56. The linear model shows an even more extreme effect: counterfactual violations drop from 49.3% to 0.0% with no change in MSE (0.5114), and ensemble variance shrinks from 19.96 to 5.73, with mean MSE improving from 8.78 to 5.64. These results confirm that PhIML can enforce philosophical constraints (either post-hoc or during training) while preserving or enhancing predictive accuracy across domains.

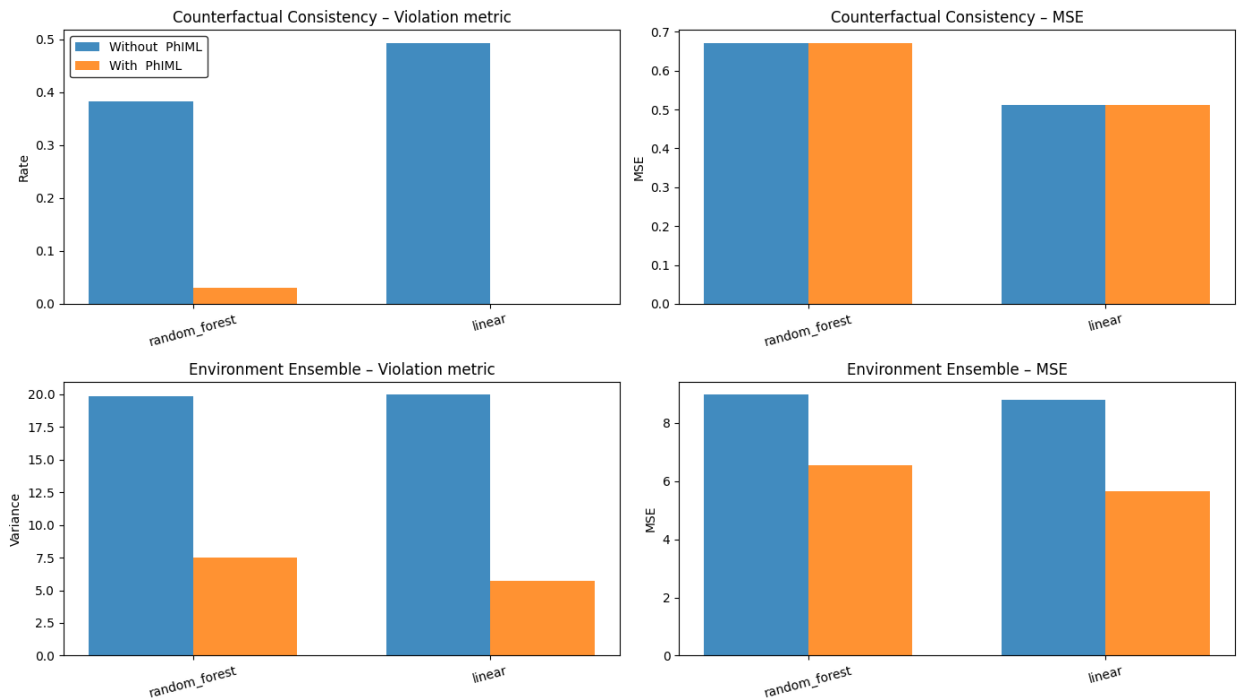

Fig. 5 Results on case study no. 2

## 4.0 Ethics & value alignment

While philosophy provides inspiration and approaches for thinking about values and ethical/moral reasoning, the integration of moral philosophy into ML represents perhaps the most challenging aspect of PhIML. This section demonstrates how philosophy can guide the development of ML systems that respect and promote human values at a conceptual level.

### 4.1 Normative theories as computational objectives

The three major traditions in normative ethics, namely *consequentialism*, *deontology*, and *virtue ethics*, each offer distinct approaches to moral reasoning. While these philosophical frameworks emerged from abstract theorizing, they can be translated into computational architectures. Among these traditions, consequentialism provides the most straightforward starting point for computational implementation. This is because its focus on outcomes aligns with ML optimization frameworks [56]. However, even the simplest utilitarian approach (i.e., maximizing aggregate welfare) immediately encounters philosophical complexities that demand sophisticated computational solutions. Consider first how *classical utilitarianism* translates into reward functions. While the basic principle of summing utilities across affected parties seems computationally trivial, this simplicity is deceptive (given the naive aggregation). For instance, the *utility monster thought experiment* demonstrates how simple summation might justify sacrificing many individuals for one being's extreme benefit [57]. This philosophical insight motivates computational approaches using bounded utility functions and weighting schemes to guarantee that no single entity's welfare can dominate.

These same theoretical concerns become practical necessities when designing reward functions [58]. For example, the challenge of *interpersonal utility comparison* (e.g., how can we meaningfully compare one person's happiness to another's?) has driven the development of preference-learning approaches [59]. Rather than assuming commensurable utility scales, these systems infer comparable metrics from revealed choices. This shift reflects a deeper philosophical move from classical hedonistic utilitarianism to preference utilitarianism, where the goal becomes satisfying desires rather than maximizing a notion of pleasure [60].

As we implement these philosophical insights computationally, additional subtleties emerge. For instance, real-world decisions involve uncertainty, which demands expected utility calculations incorporating risk attitudes and ambiguity aversion [61]. The temporal dimension adds another layer of complexity: how should we discount future welfare, and what obligations do we have to potential future people whose very existence depends on our choices? [62]. These challenges have pushed computational approaches beyond simple scalar rewards toward rich, structured value representations. Therefore, one can imply that modern systems can be designed to decompose rewards into distinct components to allow for a more nuanced moral reasoning that better reflects philosophical understanding.

Where utilitarian approaches optimize and balance, deontological ethics require absolute boundaries [63]. This fundamental difference in moral reasoning translates into a distinct computational architecture [64]. Rather than maximizing weighted sums, deontological systems enforce inviolable constraints by noting how certain actions remain forbidden regardless of their consequences. For example, implementing *Kantian ethics* computationally requires formalizing abstract principles into logical constraints and rules that remain meaningful across diverse contexts. This translation process reveals the complexity hidden within seemingly simple prohibitions, *never lie* sounds straightforward, yet computational implementation must grapple with what constitutes deception and whether context ever permits exceptions.

The architecture of deontological systems can reflect these philosophical nuances through hierarchical constraint structures. At the highest level sit absolute prohibitions derived from core principles. Below these, prima facie duties can conflict and require resolution. This hierarchy mirrors philosophical debates about moral rules' nature: are they truly absolute, or do they admit of exceptions? Particularly challenging is the implementation of universalizability checks [65]. Before taking an action, a deontological system must verify that the underlying principle can coherently apply to all agents in similar situations. This requires evaluating the immediate action and reasoning about hypothetical worlds where everyone follows the same rule. Such reasoning pushes the boundaries of current computational capabilities while remaining essential to genuine moral reasoning [66].

The third major tradition, virtue ethics, presents perhaps the most radical departure from conventional computational thinking due to its focus on character development (i.e., that good actions flow from good character rather than calculation). This Aristotelian insight requires rethinking how we structure learning systems. Computationally, virtue ethics translates into policy priors that shape learning toward stable, beneficial traits. Rather than allowing arbitrary policies that maximize reward, virtue-informed systems develop consistent dispositions. An honest agent does not calculate when honesty pays; it defaults to truthfulness as a character trait. This approach employs hierarchical Bayesian models where high-level virtues act as priors to constrain lower-level policy learning.

## 4.2 Applied philosophical frameworks

The translation of abstract moral theory into algorithmic forms reveals philosophy's practical value for ML. Rather than remaining as theoretical speculation, philosophical frameworks provide specific technical solutions to ethical challenges that can be applied across diverse ML problems [67]. Consider first how algorithmic fairness has evolved beyond naive statistical measures through engagement with political philosophy [68]. For example, Rawls's theory of justice offers a sophisticated framework that addresses fundamental questions about fair distribution that purely mathematical approaches cannot answer. When we implement Rawlsian principles computationally, the *veil of ignorance* transforms from a thought experiment to an optimization procedure [69]. Thus, by designing algorithms that maximize the welfare of the worst-off groups, we operationalize Rawls's insight that just institutions are those we would choose without knowing our position in society (as seen in the upcoming case study no. 3). This position guides the development of minimax objectives that ensure ML systems perform well across all demographic groups, particularly the disadvantaged [70].

More subtly, Rawls's *difference principle* (permitting inequalities only when they benefit the least advantaged) provides principled criteria for navigating fairness trade-offs. For example, when a medical diagnostic system performs better for certain populations, Rawlsian analysis asks whether this disparity emerges from interventions that ultimately help worse-off communities access better care. This analysis requires the model to reason counterfactually about how its deployment affects different groups [71]. Yet, Rawlsian approaches focus primarily on distributive outcomes. This limitation motivates the incorporation of Sen's capabilities approach, which shifts attention from resources or outcomes to what people can actually achieve (i.e., their capability sets). This seemingly abstract distinction has implications for ML system design. For example, a hiring algorithm informed by capabilities theory moves beyond comparing qualification scores to considering how social circumstances shaped individuals' opportunities to develop those qualifications [72].

The multidimensional nature of capabilities presents both challenges and opportunities. Unlike simple metrics that can be easily optimized, capabilities resist reduction to scalars [73]. Rather than being a limitation, this complexity prevents the oversimplification of many fairness interventions to encode ML with richer representations that respect value plurality while remaining computationally tractable [74]. This multidimensional representation enables models to explain their decisions in terms meaningful to humans, not just *this candidate scored 0.87* but *this candidate has developed capabilities in technical problem-solving despite limited access to formal education, suggesting high potential for growth*.

*4.3 Case study no. 3*

As mentioned in footnote no. 1, this experiment evaluates how integrating *Rawlsian principles* of justice into ML algorithms can mitigate entrenched biases in hiring decisions through post-hoc calibration and intrinsic (in-training) modifications. We demonstrate the merit of PhIML by comparing five distinct approaches: standard RF, post-hoc Rawlsian-enhanced RF, in-training Rawlsian RF, standard neural network (NN), and Rawlsian-aware NN in scenario no. 7. To conduct this experiment, we generate a controlled synthetic dataset of 1,500 candidates with features $x_i \in R^{\boxed{5}}$ (experience, education, test scores, skills, internship) and sensitive attributes $s_i = (\text{gender}_i, \text{ethnicity}_i, \text{socioeconomic status (SES}_i))$. We also impose systematic biases through group-specific penalties $\delta(s_i)$ on hiring scores to create single- and intersectional disadvantages that mirror real-world discrimination patterns.

The experiment implements five scenarios that progressively incorporate Rawlsian principles. The Standard RF and Standard NN serve as baselines, as they are trained solely on $\{(x_i, y_i)\}$ without considering group membership during prediction. The Post-hoc Rawlsian RF maintains the original RF model but introduces a validation-based threshold calibration mechanism that identifies the worst-performing third of demographic groups $\mathcal{G}_w$ and optimizes group-specific thresholds $\{\tau_g\}$ to maximize $\min_{g \in \mathcal{G}_w} a_g(\tau)$ while constraining overall accuracy loss to 10%. As one can see, the Post-hoc Rawlsian RF preserves the veil of ignorance during training but lifts it during validation to identify the worst-performing third of demographic groups.

For the post-hoc calibration, the veil of ignorance is strictly maintained during model training $f_0$ learns parameters $\theta$ using only $\mathcal{D}_{train} = \{(x_i, y_i)\}_{i=1}^{n}$ with no knowledge of $s_i$. Only after training completes do we lift the veil on a validation set to compute group-specific accuracies $a_g^0 = |D_g|^{-1} \sum_{i \in D_g} \mathbb{1}[f_0(x_i)=y_i]$ where $D_g = \{i: s_i \in g\}$. This temporal separation ensures the model's core parameters remain unbiased by group membership. The worst-off groups $\mathcal{G}_w$ are identified as those in the bottom tertile where $|D_g| \geq 20$. The recalibrated predictor $f_\tau(x_i, s_i) = \mathbb{1}[p_i > \tau_{g_i}]$ uses thresholds obtained by solving:

$$\max_{\{\tau_g\}} \left[ 0.7 \min_{g \in \mathcal{G}_w} a_g(\tau) + 0.3 \frac{1}{|\mathcal{G}_w|} \sum_{g \in \mathcal{G}_w} a_g(\tau) \right] \quad \text{s.t.} \quad a_{\text{all}}(\tau) \geq 0.9 \cdot a_{\text{all}}^0 \tau g \tag{13}$$

This approach operationalizes both the veil (through training ignorance) and the *difference principle* (through worst-group optimization) while maintaining aggregate performance through the constraint.

For the Rawlsian RF, splits are evaluated on features $x$ alone, but the impurity criterion incorporates group membership: given groups $\mathcal{G} = \{g_1, ..., g_k\}$ at a node, we compute group-specific Gini indices $I_{g_j} = 1 - \sum_c p_{cj}^2$ where $p_{cj}$ is the proportion of class cc c in group $g_j$. The Rawlsian impurity:

$$I_R = (1-\lambda)I_{\text{Gini}} + \lambda \left[ 0.7 \max_j I_{g_j} + 0.3 \overline{I_g} \right] \tag{14}$$

guides splitting without directly using $s$ as a splitting variable. Similarly, the Rawlsian NN maintains the veil by using $s$ only to partition the loss: forward propagation uses $\widehat{y}_i = f_\theta(x_i)$ exclusively, while backpropagation optimizes:

$$\mathcal{L}_{\mathcal{R}} = \lambda \cdot \psi\left(\{\mathcal{L}_g\}\right) + (1 - \lambda)\mathcal{L}_{BCE} \text{ where } \mathcal{L}_g = E_{i \in D_g}\left[\mathcal{L}(y_i, f_\theta(x_i))\right] \tag{15}$$

This ensures the network learns representations based on merit ($x$) while the gradient signal encourages equitable performance across groups to implement Rawls's vision of decisions made in partial ignorance of one's ultimate position. It is worth noting that the data used in this case study can be seen in Fig. 6, and the settings for each of the aforementioned algorithms are listed in Table 1.

Table 1 Algorithm configuration settings[2]

| Parameter | Standard RF | Post-hoc Rawlsian RF | In-training Rawlsian RF | Standard NN | Rawlsian NN |
|---|---|---|---|---|---|
| Model Architecture | | | | | |
| n_estimators | 100 | 100 | 100 | - | - |
| max_depth | 10 | 10 | 10 | - | - |
| hidden_dim | - | - | - | 64 | 64 |
| hidden_layers | - | - | - | 3 | 3 |
| dropout_rate | - | - | - | 0.2 | 0.2 |
| Training Parameters | | | | | |
| n_epochs | - | - | - | 100 | 100 |
| learning_rate | - | - | - | 0.001 | 0.001 |
| random_state | 42 | 42 | 42 | 42 | 42 |
| optimizer | - | - | - | Adam | Adam |
| Fairness Parameters | | | | | |
| min_group_size | - | 20 | 20 | - | 20 |
| validation_split | - | 0.20 | - | - | - |
| min_accuracy_retention | - | 0.90 | - | - | - |
| worst_off_fraction | - | 1/3 | - | - | - |
| max_worst_off_groups | - | 5 | - | - | - |
| threshold_step | - | 0.02 | - | - | - |
| use_per_group_threshold | - | False | - | - | - |
| Rawlsian Weights | | | | | |
| rawlsian_weight (λ) | - | - | 0.3 | - | 0.7 |
| minimax_weight | - | 0.70 | 0.7 | - | 0.5 |
| average_weight | - | 0.30 | 0.3 | - | 0.3 |
| variance_weight | - | - | - | - | 0.2 |
| Loss Function | | | | | |
| base_loss | - | - | - | BCE | BCE |
| reduction | - | - | - | mean | none |

---

[2] As mentioned in the first footnote, no effort was made to fully finetune the hyperparameters in the NNs and based on a simple finetuning study, that is not presented herein, it is possible to achieve some minor improvement in the performance of NNs.

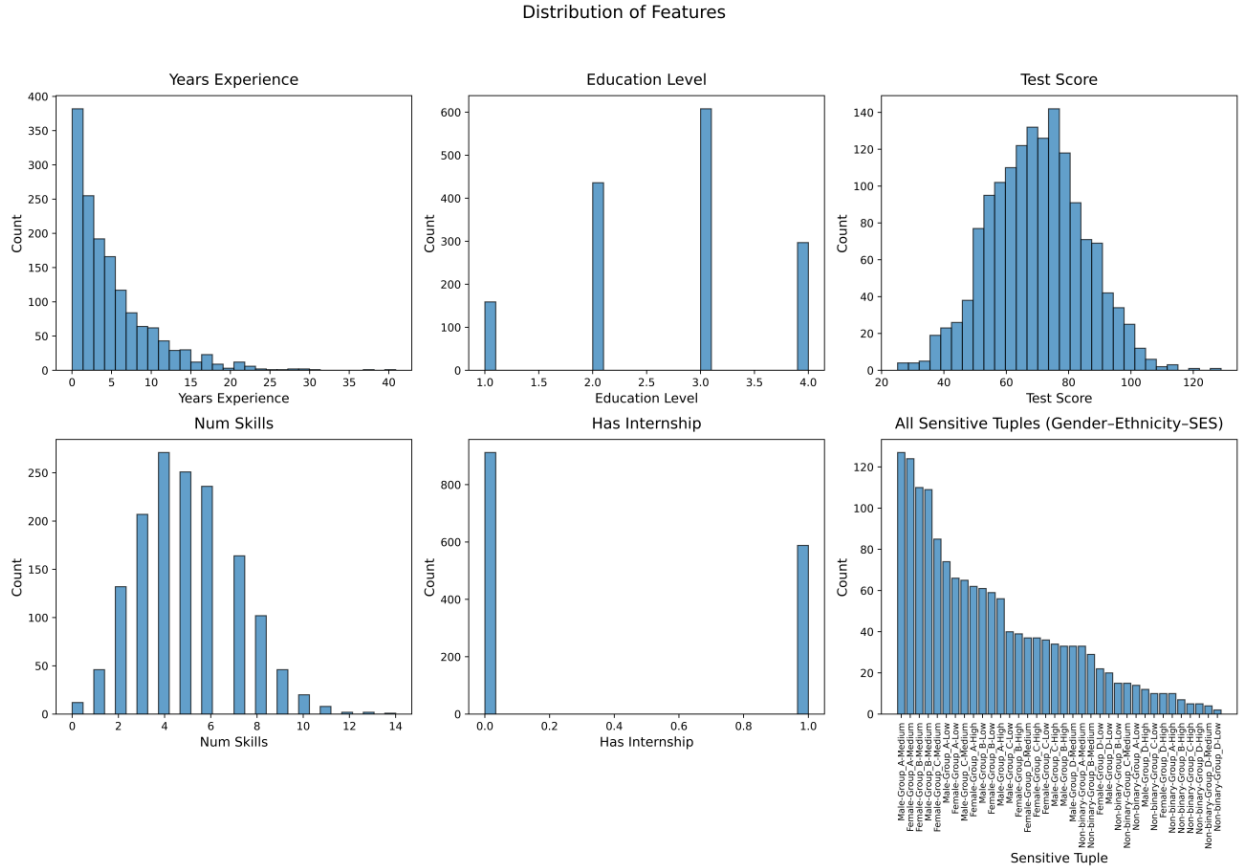

Fig. 6 Data and features used in case study no. 3

Across our biased hiring simulation, the PhIML calibration achieved clear gains with minimal sacrifice in raw performance. The experimental results reveal compelling evidence for the effectiveness of post-hoc Rawlsian calibration in achieving fairness and in improving overall performance (see Fig. 7). For example, all of the Rawlsian-related models RF achieved comparable accuracy to the standard models (with 1-2%), and while dramatically reducing disparity on at least two of the three sensitive features, the Rawls RF-in training reduced disparity on all three sensitive features.

In the synthetic data generated for this experiment, traditionally marginalized cohorts such as Non-binary applicants, Group_D ethnicity, women, and low-SES individuals receive large negative penalties in the ground-truth hiring score. Consequently, the standard RF learns that *reject* is almost always correct for them; it therefore attains strikingly high accuracies (e.g., 83.6% for women and 82.5% for Non-binary individuals) even though their predicted hiring rates hover around 20-23% (i.e., below the global 30% baseline). Conversely, males, who suffer no penalty in the generator, sit near the decision boundary, so the same RF is uncertain and attains the lowest group accuracy (61.8%).

Fig. 7a presents a two-panel comparison across all five models. The left panel contrasts overall accuracy against worst-group accuracy and shows that the performance of the standard RF and standard NN both achieve 74-75% overall accuracy (with 61.8% for their worst-performing groups). The Rawlsian variants show varied success: Post-hoc Rawls RF sacrifices minimal overall accuracy (73.3%) while lifting worst-group performance to 63.9%, whereas Rawlsian NN achieves similar worst-group gains (64.4%) with the same overall accuracy trade-off. Surprisingly, the in-training Rawls RF maintains high overall accuracy (75.1%) and achieves the best worst-group performance (65.4%), though as we show below, this does not translate to fairer hiring decisions.

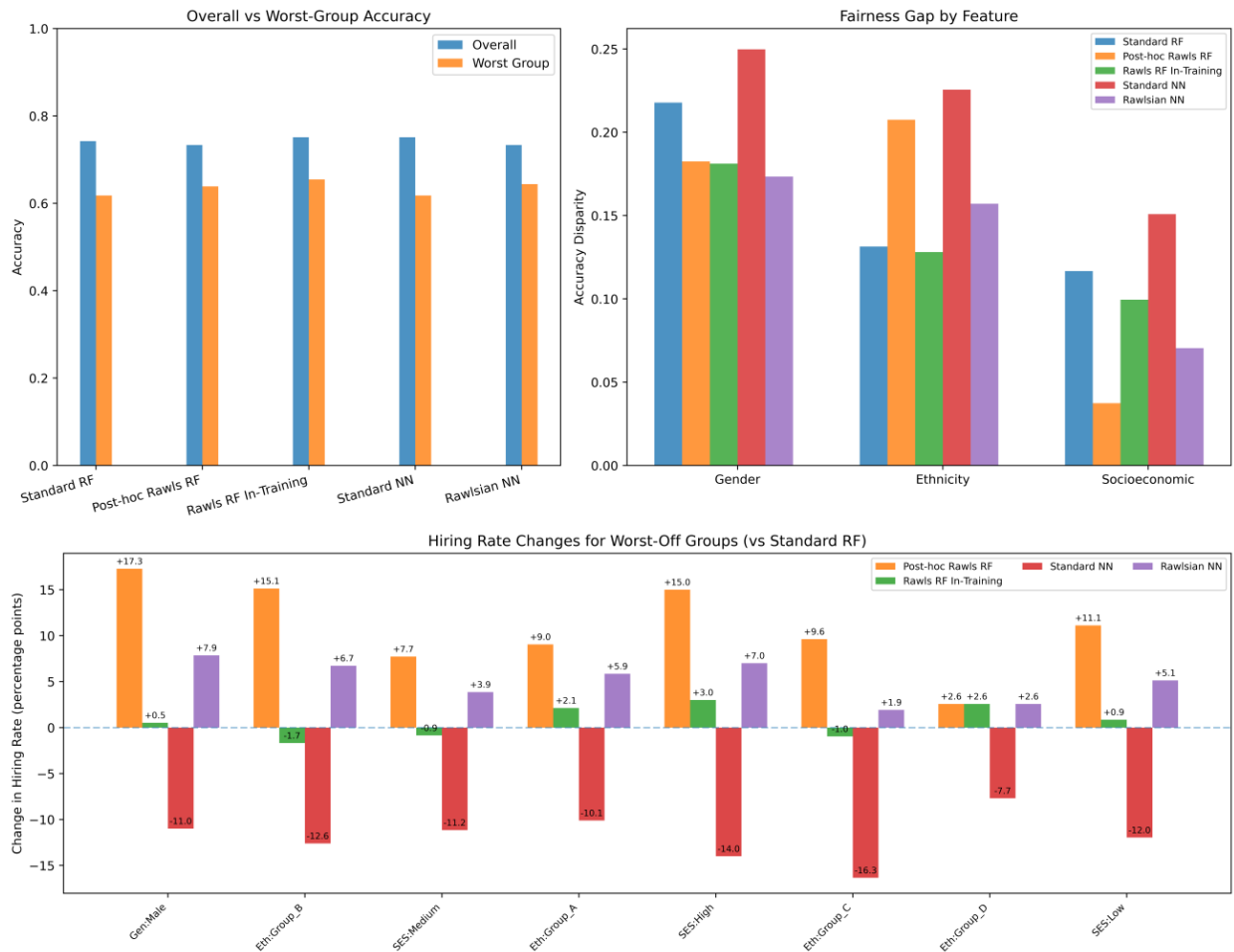

Fig. 7 Results on case study no. 3

The right panel quantifies accuracy disparity (i.e., the gap between best and worst performing groups) for each sensitive feature. Standard NN exhibits the largest disparities across all dimensions, while the Rawlsian models compress these gaps substantially. Post-hoc Rawls RF shows an interesting pattern, wherein reducing gender disparity to 18.3 points and socioeconomic disparity to just 3.7 points, it slightly increases ethnicity disparity to 20.7 points. This may suggest that uniform threshold adjustments can have heterogeneous effects across different protected attributes.

Then, we isolate the eight groups with the highest and lowest baseline accuracies by overlaying their hiring rates across all models relative to the standard RF. Post-hoc Rawls RF dominates with improvements. The bar labels reveal that males gain 17.3 percentage points, while Group_B ethnicity gains 15.1 points. Rawlsian NN shows consistent but smaller gains (6-8pp), while standard NN displays negative changes for most groups, which confirms its existing biases. The in-training Rawls RF hovers near zero and demonstrates that modifying the tree-building process, as carried out here, does not seem to effectively translate to fairer predictions.

Overall hiring rates reveal the baseline problem: standard RF assigns positive predictions to only 23.2% of applicants despite the 30% design target, with remarkably similar rates for worst-off (22.5%) and best-off (23.5%) groups—a deceptive equality masking severe under-representation relative to the true positive rate. Post-hoc Rawls RF raises the overall rate to 34.0%, with worst-off groups actually slightly exceeding best-off groups (33.4% vs 32.9%), while maintaining reasonable overall accuracy (73.3%). The improvement metrics tell the clearest story. Post-hoc Rawls RF achieves a 48.7% increase in worst-off hiring rates and reduces the equity gap by 151.2%, reversing the gap so that previously disadvantaged groups now have slightly higher rates. Rawlsian NN shows meaningful but smaller gains (22.8% worst-off improvement, 116.1% gap reduction), while in-training Rawls RF achieves negligible 3.1% improvement despite its superior accuracy metrics.

Taken together, these results make a coherent argument about PhIML, with post-hoc calibration being the most reliable operationalization of Rawls's difference principle in this hiring context. This is due to the fact that this model works directly on calibrated probabilities, and it can reallocate positive decisions with precision. The approach successfully identified that a threshold of approximately 0.30 maximizes the minimum group performance while maintaining 90% of baseline accuracy. On the other hand, the in-training approaches depend delicately on hyperparameters and architectural choices. The in-training Rawls RF's failure to improve hiring equity despite better accuracy metrics suggests that other tree-growing heuristics may overwhelm fairness-aware impurity measures. As seen above, the Rawlsian NN shows more promise, which may indicate that direct optimization of fairness objectives in neural architectures can yield meaningful, if moderate, improvements.

The synthetic experiment therefore reinforces the practical merit of PhIML's *measure-then-tune* ethos: diagnose where the baseline model embeds structural bias, then let a transparent optimization step correct the decision boundary rather than hoping intrinsic learning dynamics will land in a fair equilibrium. The 151.2% gap reduction achieved by post-hoc adjustment—essentially eliminating and slightly reversing historical disadvantage—demonstrates that simple, interpretable interventions can be remarkably effective when grounded in coherent philosophical principles.

## 5.0 Challenges & outlook
The progress in PhIML confronts fundamental challenges that span technical, philosophical, and practical dimensions. This section examines some of such obstacles.

### *5.1 Technical challenges*
The integration of symbolic and numeric computation encounters scalability barriers that limit philosophy-informed approaches to relatively small problems (compared to pure ML methods) [75,76]. For instance, grounding first-order logic requires considering exponentially many possible variable assignments, and even restricted fragments quickly become intractable for realistic problem sizes [77]. While current implementations handle domains with hundreds of entities and logical constraints, real-world applications often involve millions of entities with complex relational structures that overwhelm symbolic reasoning components [78]. The recent advances in approximate logical reasoning provide partial solutions, but the reader is to note that such reasoning also introduces new challenges around approximation quality. For example, Markov logic networks can achieve better scaling by sacrificing completeness for tractability [44]. However, these approximations may violate logical properties (such as a safety constraint that should hold universally might be satisfied only probabilistically, which undermines the guarantees that motivated logical integration). Thus, one challenge lies in developing approximation schemes that preserve essential logical properties while achieving efficiency [79].

Further, PhIML is likely to encounter undecidability [80]. In essence, modal epistemic logics with nested knowledge operators can become undecidable at depths that real-world scenarios routinely exceed. This creates an unbridgeable gap between philosophical frameworks that assume unbounded introspection and limited computational systems. One particular example can be seen in ethical frameworks that consider infinite populations or unbounded causal chains [81]. Practically, rather than treating these results as defeats, interested researchers can aim to develop principled restrictions that maintain decidability while preserving essential reasoning capabilities (e.g., incorporate depth-bounded logics, limit knowledge operator nesting to cognitively plausible levels, etc.). These also reflect a core belief: any real-world intelligence must operate within computational limits that shape how it thinks and what it can think about.

## *5.2 Philosophical challenges*

The formalization of philosophical concepts into precise computational frameworks faces fundamental tensions between conceptual richness and formal precision [82]. For example, concepts like human dignity and moral agency carry centuries of philosophical analysis that resist reduction to simple formal definitions [83]. Therefore, attempts to operationalize dignity through preference satisfaction or capability approaches capture some aspects while missing others that philosophers consider essential. This challenge is not only technical but conceptual—how can computational systems respect human dignity when philosophers themselves debate its meaning? [84].

Similarly, cross-cultural variation in ethical frameworks also poses challenges for philosophy-informed systems intended for global deployment [85]. This is because virtue ethics emphasizes different virtues across cultures (e.g., individualistic societies prioritize autonomy and self-reliance while collectivistic cultures emphasize harmony and reciprocal obligation, etc.) [86]. In parallel, deontological principles vary, and it becomes difficult to identify what constitutes fair treatment, justified deception, or appropriate authority across cultural contexts. There is also the challenge of incompatible ethical theories providing conflicting guidance (e.g., utilitarian calculations sometimes demand actions that violate deontological constraints, while virtue ethics may recommend character traits that reduce aggregate welfare). While simplistic approaches often aggregate multiple ethical theories through some form of voting/weighing, this risks incoherent compromise positions that satisfy no ethical framework fully. Therefore, philosophy-informed systems should be designed to navigate ethical pluralism without defaulting to relativism or imposing particular cultural values.

As the above case studies show, each experiment isolates a single philosophical principle, applied either entirely post-hoc or exclusively within model training, rather than combining multiple principles or layering intrinsic and post-hoc treatments simultaneously. This one-at-a-time approach prevents unintended interactions between constructs, but it also limits our ability to assess synergy or conflict among co-occurring philosophical interventions. Addressing combinations of PhIML concepts, whether multiple post-hoc adjustments, multiple in-training modifications, or hybrid pipelines that integrate both, will require careful strategy design, new evaluation metrics, and potentially novel optimization frameworks. Such multifaceted applications fall outside the present scope but represent a rich avenue for future research.

### *5.3 Practical challenges*

The industrial adoption of PhIML approaches faces a number of practical barriers: 1) technical complexity, 2) infrastructure lock-in, and 3) organizational inertia. For example, companies that have invested in specialized infrastructure optimized for traditional ML's GPU/CPU could suffer from substantial switching costs when considering systems that require specialized hardware for logical reasoning [87]. Along the same lines, incorporating formal epistemological pipelines can extend development cycles, which may not be looked upon favorably in a speed-driven deployment field.

Other practical challenges revolve around human capital and governance, especially in terms of inadequacies in educating technologists and regulating intelligent systems [88]. For example, the interdisciplinary expertise required for fluency in modal logic, category theory, ML, and normative ethics exists at the intersection of fields that have evolved increasingly divergent methodological cultures/technical languages. In a way, training professionals who can navigate such fields requires not simply adding philosophy courses to computer science curricula or programming workshops to philosophy departments, but fundamentally reconceptualizing how we train researchers to address multi-domain questions [89]. The same implies that current regulatory approaches that focus on post-hoc auditing or mechanical compliance with data protection rules lack the conceptual sophistication to evaluate whether an ML system's formal representation of human agency respects philosophical commitments to autonomy [90]. Other questions worth exploring include: how should regulators evaluate whether a system properly implements informed consent when using formal theories of agency? What standards should govern the selection of ethical frameworks for value alignment? How can democratic input shape the philosophical commitments embedded in ML systems?

## 6.0 Conclusions

Philosophy-informed ML (PhIML) arises as more than an interdisciplinary curiosity that represents a necessary growth in how we design, train, and deploy ML systems. This is because the integration of epistemology, logic, causation, and ethics into ML addresses fundamental limitations that purely statistical approaches cannot overcome. In particular, this work notes how PhIML will likely enhance rather than hinder the practical capabilities of ML systems. While significant challenges remain, the field's trajectory points toward ML systems worthy of the trust society increasingly places in them. Thus, as ML systems assume greater autonomy and influence, the question is not whether to incorporate philosophical principles but how quickly we can develop the frameworks, tools, and talent to do so effectively. Other key findings include:
- PhIML aims to address critical limitations of current ML, namely: blackbox brittleness (through epistemologically-grounded uncertainty representation), causal blindness (through formal theories of causation), and alignment failures (through multi-framework ethical integration).

- Philosophical foundations enable superior performance in limited data regimes where statistical methods fail. This is because epistemic priors provide sophisticated inductive biases.
- The interdisciplinary nature of PhIML demands new educational approaches and governance frameworks that can evaluate philosophical adequacy alongside technical performance.

## Data availability
None.

## Conflict of interest
The author declares no conflict of interest.

## Funding
None.

## Authors' contributions
None.

## Acknowledgements
None.